# Ontology-supported processing of clinical text using medical knowledge integration for multi-label classification of diagnosis coding

Phanu Waraporn[1,4,*], Phayung Meesad[2], Gareth Clayton[3]
[1]Department of Information Technology, Faculty of Information Technology
[2]Department of Teacher Training in Electrical Engineering, Faculty of Technical Education
[3]Department of Applied Statistics, Faculty of Applied Science
King Mongkut's University of Technology North Bangkok
[4]Division of Business Computing, Faculty of Management Science
Suan Sunandha Rajabhat University
Bangkok, Thailand
[*]phanu.waraporn@gmail.com, pym@kmutnb.ac.th , gareth@kmutnb.ac.th

*Abstract*—**This paper discusses the knowledge integration of clinical information extracted from distributed medical ontology in order to ameliorate a machine learning-based multi-label coding assignment system. The proposed approach is implemented using a decision tree based cascade hierarchical technique on the university hospital data for patients with Coronary Heart Disease (CHD). The preliminary results obtained show a satisfactory finding.**

*Keywords-component; medical ontology, diagnosis coding, knowledge integration, machine learing, decision tree.*

## I. INTRODUCTION

From the fact that the typical character of any medical data is heterogeneous, traditional machine learning approach alone cannot be directly applied to solve efficiently our study in the area of automating the clinical coding task. Therefore, we present a knowledge integration method based on the utilization of distributed medical ontology support knowledge capturing and integration and machine learning techniques to enhance a coding assignment of multi-label medical text.

Text Categorization or Text Assignment (TA) as part of the Natural Language Processing (NLP) consists the assignment of one or more preexisting categories to a text document [1]. In multi-label assignment, the problem can comprise various classes.

As large unstructured and structured medical databases are being generated momentarily, difficulties accessing, integrating, extracting, and managing knowledge out of them are among many reasons researchers are trying to overcome including utilizing ontology, a form of knowledge-based systems which are repositories of structured knowledge such as UMLS, MeSH, ICD, etc.

According to Nelson et al.[2], several studies have shown that the use and integration of several knowledge sources improves the quality and efficiency of information systems using the query on the ontology , specifically so in the domain specific such as the health information systems or medicine. An ontology is a specification of a conceptualization that defines and/or specifies the concepts, relationships, and other distinctions that are relevant for modeling a domain. Such specification takes the form of the definitions of representational vocabulary (classes, relations, and so on), which provide meanings to the vocabulary and formal constraints on its coherent use [3].

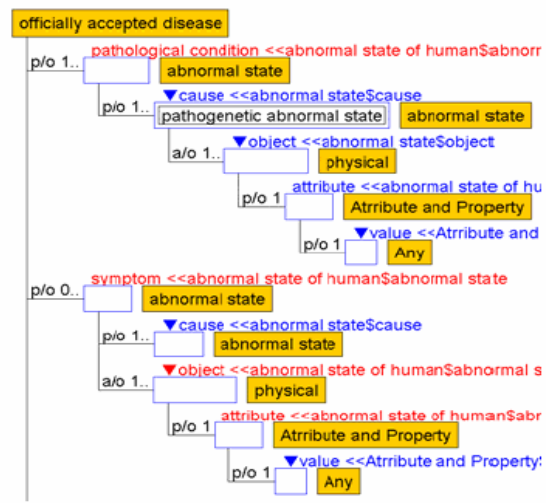

Fig.1. The framework of officially accepted disease [6].

In compliance with the WHO ICD 10 for Coronary Heart Disease [4], a domain specific ontology based on a distributed architecture is constructed for use in the work to support the knowledge capturing and integration processes. As depicted in Figure 1 below an adapted framework of officially accepted disease generated in HOZO [5] representing a high level medical ontology in our work which can be shared and reused.

The essence in managing knowledge about semantic relations, how to acquire as well as how to integrate them and finally how to put them altogether are discussed. Of many machine learning algorithms, we short-listed few of them and preliminary tests are performed. We compare the results with our own classifier, a variant of the existing decision tree, and conclude that our approach marginally outperform the traditional ones.

The rest of the paper is arranged as follows: first, a related work section comments on some relevant works in the field; section 3 briefly introduces the proposed system architecture; section 4 summarizes the data collection and experiments along with the preliminary results. We deliver the conclusions and future work in the epilogue of section 5.

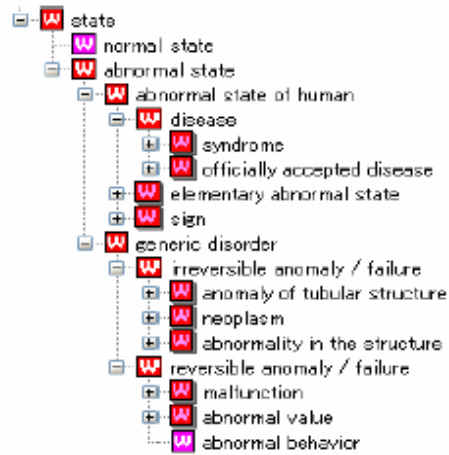

Fig.3  High Level category in HOZO [6].

## II.  RELATED WORKS

*A. Ontology*

Our aim of enhancing the automatic assignment of medical coding by automatically integrating ontologies is guided by the use of ontologies in various natural language processing tasks such as automatic summarization, text annotation and word sense disambiguation, among other [7]. Advantage of using ontologies in the area of relevance-feedback, corpus-dependent knowledge models and corpus-independent knowledge models on the domain-specific and domain-independent ontologies all contribute to ameliorate information retrieval systems [8].

Domain-independent ontologies such as WordNet/ Medical WordNet though improves a word sense disambiguation, it has so broad coverage that it can be debatable for the ambiguous terms making a domain-specific ontologies, particularly on the part of a terminology which is less ambiguous. Further more, it models terms and concepts corresponding to a specific or given domain [9].

In our case, an Ischaemic Heart Disease ontology is manually built by reusing the framework, high level, relations and concepts defined in accordance with [6]. Figures 2 and Figure 3 illustrates the property inheritance and top level, respectively.

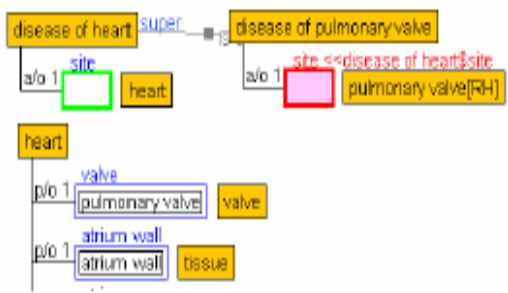

Fig.2  Property Inheritance in HOZO Heart Disease [6].

*B. Medical Knowledge Integration Framework and Process*

In a complex domain such medicine, knowledge integration requires consolidation of the heterogeneous knowledge and reconfirm it with the induced models since it originates from sources with different levels of certainty and completeness. Therefore, new models are collectively learned and comprehensively evaluated based on existing knowledge. Figure 4 illustrates the overview of medical knowledge integration whilst a more comprehensive medical knowledge integration process is depicted in Figure 5.

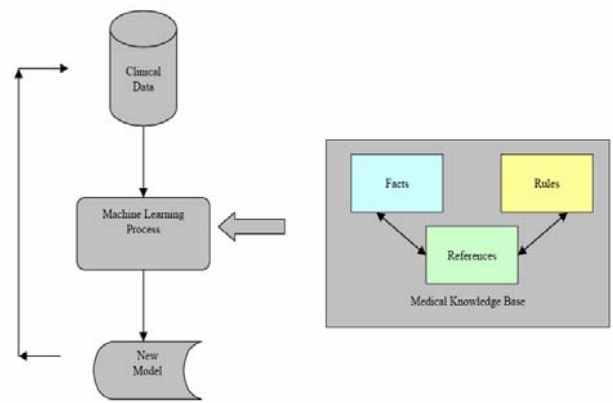

Fig.4  Overview of medical knowledge integration architecture, adapted from [10].

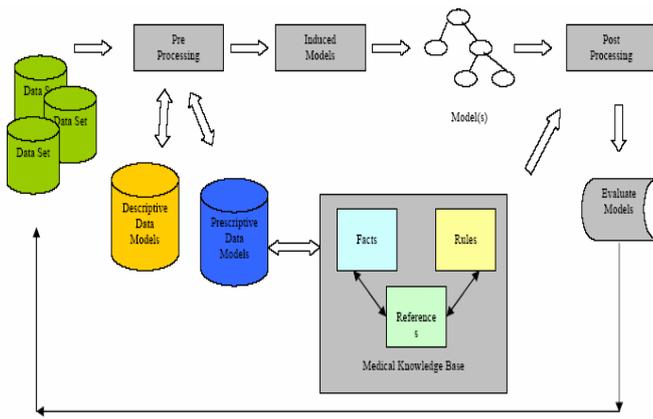

Fig.5 Domain knowledge integrated into medical knowledge integration process, adapted from [10].

*C. Clinical data/text*

Patient records in use are the discharge summary records generally contain the principal diagnosis (PDx) information. The particular of secondary diagnosis (SDx) may appear provided that there is additional ailment requiring treatment in addition to the PDx. If any, SDx can be the details about the comorbidity, complication, status of the disorder; chronic, acute and external cause of the injury. Additionally, depending on the necessity, the treatment may involve other medical procedures (PROC) and such details are to be itemized. In the case where only PDx is reported, the coding process is fairly simple and fast. Generally, however, the patient records contain many SDxs and multiple PROC. The coder duty is to translate those PDx(s), SDx(s) and PROC information into corresponding ICD 10 code(s). Other patient data aside from the discharge summary includes but not limited to prescription information sheet, laboratory results, drug prescribed nurse note, progress note, and other records [11].

*D. Multi-label Classification Task[12][13]*

A pattern recognition, classification, categorization can be viewed as a mapping from a set of input variables to an output variable representing the class label. In classification problems the task is to assign new inputs to labels of classes or categories. As input data the patient's anamnesis, subjective symptoms, observed symptoms and syndromes, measured values (e.g. blood pressure, body temperature etc.) and results of laboratory tests are taken. This data is coded by vector x, the components of which are binary or real numbers. The patients are classified into categories $D_1, \ldots, D_m$ that correspond to their possible diagnoses $d_1, \ldots, d_m$. Also prediction of the patient's state can be stated as classification problem. On the basis of examined data represented with vector x patients are categorized into several categories $P_1, \ldots, P_m$ that correspond to different future states. For example five categories with the following meaning can be considered: $P_1$ can mean death, $P_2$ deterioration of the patient's state, $P_3$ steady state, $P_4$ improvement of the patient's state and $P_5$ recovery.

As opposed to binary classification, multiple labels (e.g., Breathing, Blood circulation, and Pain) are considered. In binary classification there was one label (e.g., Breathing) and the task was to assign instances either to the class of topically related (i.e., belongs to the class of Breathing) or to the class of topically unrelated (i.e., does not belong to the class of Breathing) objects. In other words, the multi-label classification task can be restructured to multiple binary classification tasks - one for each topic-label. The only difference in the multi-class classification is that in multi-label classification, each instance can belong to several classes at the same time.

*E. Diagnosis Coding*

From Figure 6, a ICD 10 for Coronary Heart Disease (CHD) is presented in a hierarchical form. The top first level is the high/concept level class where the second and third levels are major and minor classes, respectively. The minor class gives more specific information yet not comprehensively enough when compared with other major medical concepts and ontologies like UMLS, GALEN, SNOMED CT, etc. This is due to the fact that information is served differently.

Key words/terms extracted from patient discharge summary will undergo the coding process and provided that all relevant information are in accordance, this summary will finally be assigned a corresponding ICD 10 code(s) as per specified in the ICD 10.

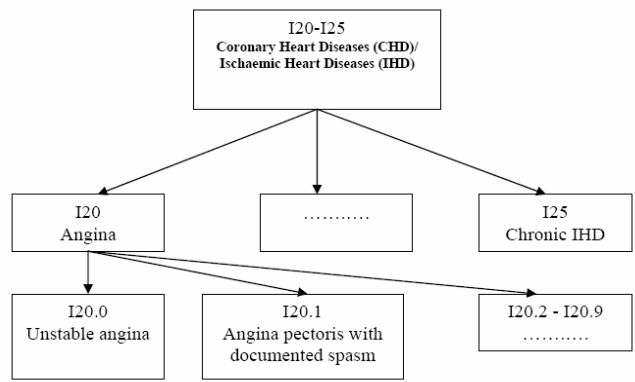

Fig.6 ICD 10 for Coronary Heart Disease in hierarchical form.

With the preceding background and related works, a system for automating the ICD 10 coding assignment for CHD can now begin. In the next section, a framework for such a system will be introduced. Furthermore, data collected for this preliminary experiment will be briefed and early result of applying the proposed novel method called cascade hierarchical Decision Tree (chiDT), a twice processing of Decision Tree C4.5 algorithm, is presented.

### III. THE PURPOSED FRAMEWORK AND PRELIMINARY EXPERIMENTAL RESULTS

This section explains the proposed framework appeared in Figure 7 below. Our emphasis is on the one surrounded by the red block.

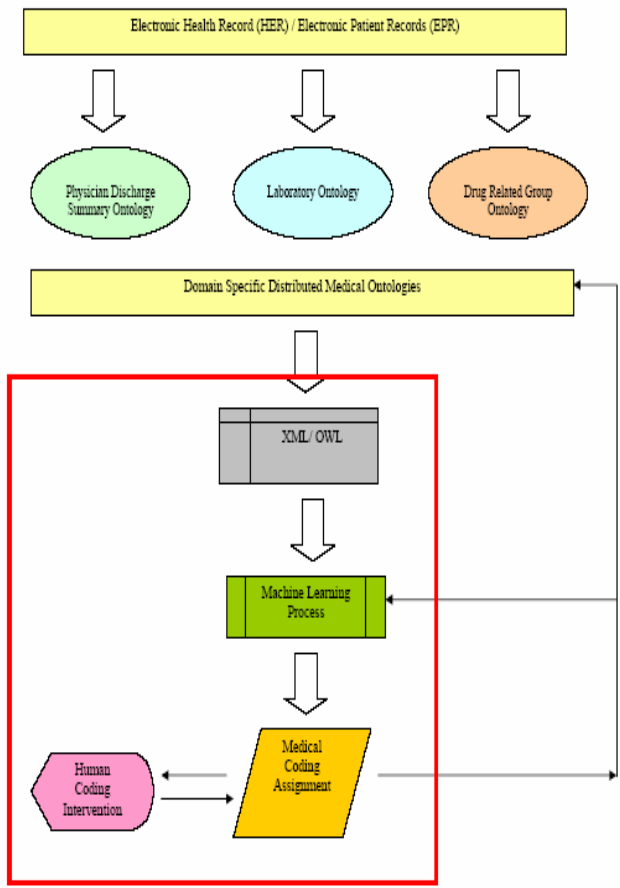

Fig.7 High level framework for automating ICD 10 coding assignment.

A machine-learning approach using a cascade of two classifiers trained on the same data is used to predict the codes. Both classifiers are trained with the same data, and perform multi-label classification by decomposing the task into 11 (out of 28) binary classification problems, one for each code. In this setting, it is possible for a classifier to predict an empty, or impossible, combination of codes. Such known errors are used to trigger the cascade: when the first classifier makes a known error, the output of the second classifier is used instead as the final prediction. No further correction of the output of the second classifier is performed as preliminary experiments suggested that this would not further improve the performance.

In the context of Thai people based on data collected during year 2009 for all patients with Coronary Heart Disease (CHD) in one university hospital, out of 28 ICD 10 codes in the CHD group, 11 classes were reported, the rest is non existed.

196 patients with CHD were reported to the university hospital during 2009. 53 patient records were then separated for training purpose. These instances were carefully selected in order to cover all the classes. In order to understand the data for preliminary testing purpose, we recombined 53 training records with the 143 remaining and untrained ones and based on chiDT classifier using WEKA [14] as a processing engine, the following early results were reported. The following tables detail the results generated for each different classifier.

Table 1 represents the results of 53 carefully selected records covering 11 classes widely seen in Thailand and used for training. While Tables 2 to 4 present the results on all 196 records used for training the chiDT, SVM and Fuzzy classifiers, respectively.

| chiDT | | |
|---|---|---|
| Correctly Classified Instances | 52 | 98.1132 % |
| Incorrectly Classified Instances | 1 | 1.8868 % |
| Kappa statistic | 0.9756 | |
| Mean absolute error | 0.012 | |
| Root mean squared error | 0.0818 | |
| Relative absolute error | 4.5652 % | |
| Root relative squared error | 22.5907 % | |
| Total Number of Instances | 53 | |

Table 1 With original 53 selected for training, the chiDT alone gave approximately 98.11 % accuracy.

| chiDT | | |
|---|---|---|
| Correctly Classified Instances | 185 | 94.3878 % |
| Incorrectly Classified Instances | 11 | 5.6122 % |
| Kappa statistic | 0.9323 | |
| Mean absolute error | 0.0107 | |
| Root mean squared error | 0.0927 | |
| Relative absolute error | 7.0466 % | |
| Root relative squared error | 33.6758 % | |
| Total Number of Instances | 196 | |

Table 2 With original 196 selected for training, the chiDT alone gave approximately 94.39 % accuracy.

| SVM | | |
|---|---|---|
| Correctly Classified Instances | 183 | 93.3673 % |
| Incorrectly Classified Instances | 13 | 6.6327 % |
| Kappa statistic | 0.9196 | |
| Mean absolute error | 0.1492 | |
| Root mean squared error | 0.2642 | |
| Relative absolute error | 97.9788 % | |
| Root relative squared error | 95.9388 % | |
| Total Number of Instances | 196 | |

Table 3 With original 196 selected for training, the chiDT with Support Vector Machine (SVM) classifier gave approximately 93.37 % accuracy.

| FLR | | |
|---|---|---|
| Correctly Classified Instances | 184 | 93.8776 % |
| Incorrectly Classified Instances | 12 | 6.1224 % |
| Kappa statistic | 0.9262 | |
| Mean absolute error | 0.0111 | |
| Root mean squared error | 0.1055 | |
| Relative absolute error | 7.3094 % | |
| Root relative squared error | 38.3185 % | |
| Total Number of Instances | 196 | |

Table 3 With original 196 selected for training, the chiDT with Fuzzy classifier gave approximately 93.87 % accuracy.

The early result of this preliminary experiments displays that the proposed chiDT classifier based on cascade hierarchical architecture does produce satisfying output followed marginally by Fuzzy and SVM techniques.

## IV. CONCLUSION AND FUTURE WORKS

In this paper, we have provided an overview of the proposed machine learning algorithm titled cascade hierarchical Decision Tree (chiDT). Though the result was not so distinctively conclusive due to limited number of the data set, the authors expect that with increased sample size, the result should improve the yield substantially. To conclude that this algorithm is reliable, a cost sensitivity analysis (CSA) will be tested apart from other standard model evaluation techniques. These works represent an upcoming work to be carried out. Also it would be appropriate to mention that by integrating ontology and medical knowledge integration framework into this work, future medical knowledge management would enhance both works of computer scientists and clinical specialists, to name just a few, for the improved classifiers and clinically generic models, respectively.

## V. ACKNOWLEDGEMENT

We would like to extend our sincere thanks to staff at the Mahidol University's Faculty of Medicine Siriraj Hospital Division of Molecular Genetics, staff at the National Center for Genetic Engineering and Biotechnology (BIOTEC) and staff at the Human Language Technology Laboratory of the National Electronics and Computer Technology Center (NECTEC) for all resources and advices discussed in this experimental study. In addition, we acknowledge the team at the Mizoguchi Laboratory of the Institute of Scientific and Industrial Research, Osaka University for allowing us to access the software, HOZO, an ontology editor.

AUTHORS PROFILE

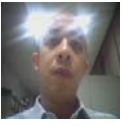

**Phanu Waraporn**, a Ph.D. candidate in Information Technology at the Faculty of Information Technology, King Mongkut's University of Technology North Bangkok. Currently, he is the lecturer in Business Computing at the Faculty of Management Science, Suan Sunandha Rajabhat University, Bangkok, Thailand.

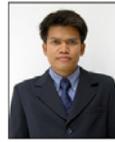

**Phayung Meesad, Ph.D.**, assistant professor. He is an associate dean for academic affairs and research at the faculty of Information Technology, King Mongkut's University of Technology North Bangkok. He earned his MS and Ph.D. degrees in Electrical Engineering from Oklahoma State University, U.S.

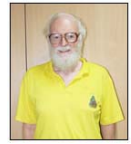

**Gareth Clayton, Ph.D.**, a senior lecturer in statistics at the Department of Applied Statistics, Faculty of Applied Science, King Mongkut's University of Technology North Bangkok. He earned his Ph.D. in Statistics from Melbourne University, Australia.